\documentclass{article}
\usepackage[sglblindworkshop, final]{neurips_2026}
\workshoptitle{AgenticOS}
\usepackage[utf8]{inputenc}
\usepackage[T1]{fontenc}
\usepackage{hyperref}
\usepackage{url}
\usepackage{booktabs}
\renewcommand{\arraystretch}{0.85}
\usepackage{amsfonts}
\usepackage{nicefrac}
\usepackage{microtype}
\usepackage{xcolor}
\usepackage{amsmath}
\usepackage{amssymb}
\usepackage{graphicx}
\usepackage{adjustbox}
\usepackage{multirow}
\usepackage{pifont}
\usepackage{enumitem}
\usepackage{float}
\usepackage{placeins}
\usepackage{caption}
\makeatletter
\renewcommand{\section}{%
  \@startsection{section}{1}{\z@}%
                {-1.6ex \@plus -0.4ex \@minus -0.2ex}%
                { 1.0ex \@plus  0.2ex \@minus  0.1ex}%
                {\large\bf\raggedright}%
}
\renewcommand{\subsection}{%
  \@startsection{subsection}{2}{\z@}%
                {-1.4ex \@plus -0.4ex \@minus -0.2ex}%
                { 0.5ex \@plus  0.1ex}%
                {\normalsize\bf\raggedright}%
}
\makeatother

\title{Kernel-Managed Shared Memory for System-Wide Personalization}

\author{%
  Ryan Lum \\
  Department of Computer Science\\
  Rutgers University\\
  Piscataway, NJ 08854 \\
  \texttt{rkl49@scarletmail.rutgers.edu}
  \And
  Yongfeng Zhang \\
  Department of Computer Science\\
  Rutgers University\\
  Piscataway, NJ 08854 \\
  \texttt{yongfeng.zhang@rutgers.edu}
}

\begin{document}
\FloatBarrier
\maketitle

\begin{abstract}
AI systems become more useful when they can adapt to the people using
them, but in multi-agent systems, useful context learned by one agent
often remains unavailable to others. We present kernel-managed shared
memory, a system-level abstraction in which specialized agents write
structured, tagged memories while the agent-system kernel -- not
individual agents -- governs retrieval, privacy enforcement, and prompt
injection. We implement and evaluate this design on AIOS and compare it
against three alternatives across three assistant models (GPT-4o,
Llama-3.1:8B, Qwen-2.5:7B) and 1,800 total trials. Against an unmanaged
external memory backend (Mem0) using identical underlying storage,
kernel-managed retrieval and injection improve personalization scores by
$2.4$--$4.0$ points on a 5-point scale (e.g., $1.05\!\to\!4.69$ profile
usage on GPT-4o), with every comparison significant at $p<10^{-18}$.
Against standard retrieval-augmented injection, gains are similarly large
and consistent across all three models. Against full, unfiltered
context concatenation -- a soft ceiling on available context rather than
on response quality -- kernel-managed injection statistically matches
performance on two of three models and shows a small, model-specific
deficit on the third, while using substantially shorter prompts:
end-to-end latency is $15$--$61\%$ lower across all three models, with
corresponding reductions in per-call token usage and inference cost.
These results indicate that centralizing memory management in the
agent-system kernel, rather than leaving retrieval and privacy
enforcement to individual agents, delivers most of the personalization
benefit of unconstrained context at a fraction of its cost.
\end{abstract}

\begin{figure}[t]
    \centering
    \includegraphics[width=0.85\linewidth]{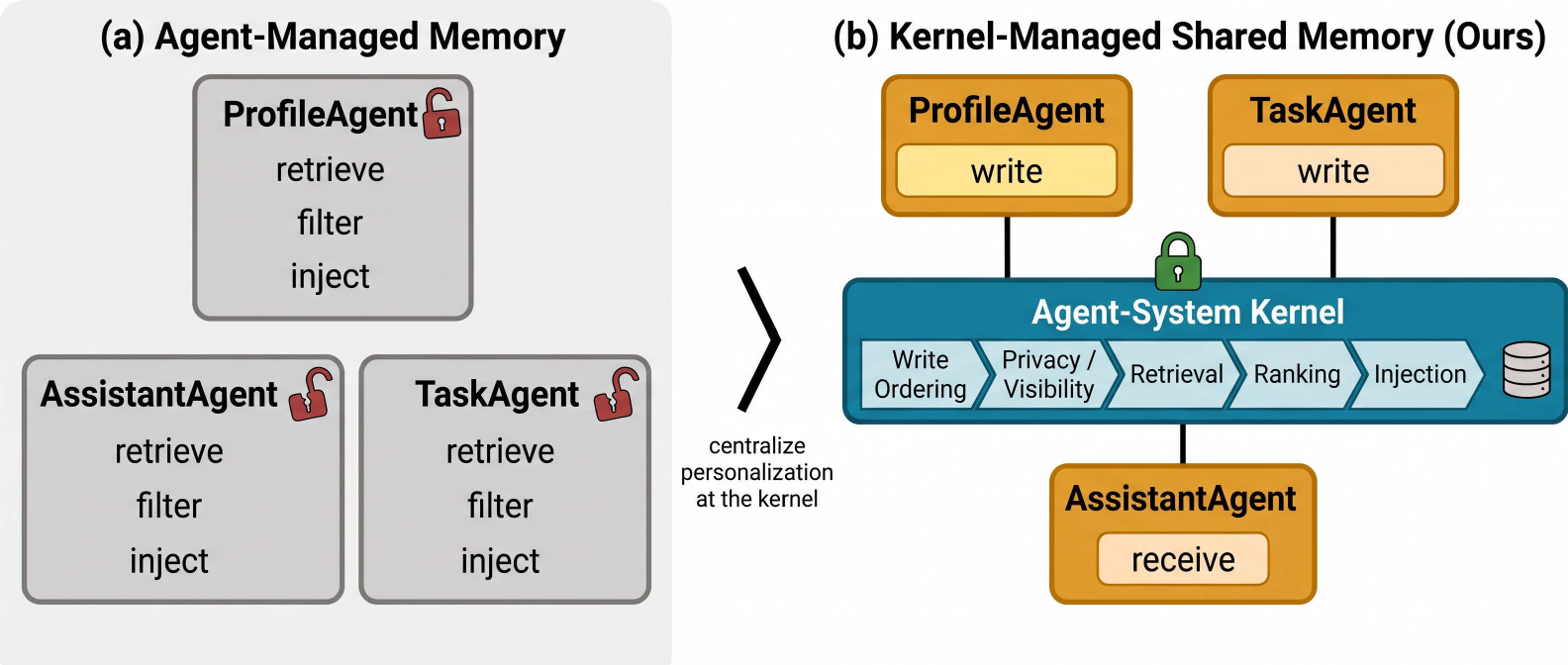}
    \caption{Kernel-managed shared memory centralizes retrieval, privacy
    enforcement, and prompt injection at the agent-system kernel instead of
    leaving them to individual agents (left vs.\ right), matching
    full-context personalization quality at a fraction of the latency and
    token cost (Section~\ref{sec:results}).}
    \label{fig:teaser}
\end{figure}

\section{Introduction}
\label{sec:intro}

Personalization is a key ingredient for making AI systems useful: a
system that remembers a user's preferences, current work, and past
interactions produces more relevant responses over time. This is
especially true in multi-agent systems, where different agents learn
different parts of a user's context -- one agent may learn stable
preferences, another current task context, a third may respond directly
to the user -- and if that information stays isolated, the system cannot
behave coherently as a whole.

Figure~\ref{fig:teaser} contrasts two ways of organizing this. Under
\emph{agent-managed memory} (Figure~\ref{fig:teaser}a), each agent
independently retrieves, filters, and injects its own memory, duplicating
logic across agents and leaving privacy enforcement to per-agent
convention -- a design that is easy to get wrong in exactly the way we
describe in Section~\ref{sec:metadata-visibility}. We instead propose
\emph{kernel-managed shared memory} (Figure~\ref{fig:teaser}b): agents
write structured, tagged memories, while a runtime layer distinct from
any individual agent -- the agent-system kernel -- centralizes write
ordering, privacy enforcement, retrieval, ranking, and injection. This
makes personalization a systems capability rather than repeated
application-level logic. We instantiate and evaluate this design on
AIOS, an existing operating-system-style framework for LLM-based agents,
though the underlying abstraction is not specific to AIOS.

We evaluate this approach against three alternatives -- an unmanaged
external memory backend, standard retrieval-augmented injection, and full
unfiltered context concatenation -- across three assistant models and
1,800 total trials. Kernel-managed shared memory substantially and
significantly outperforms the unmanaged memory backend and standard
retrieval-augmented injection on every model tested. Against full
unfiltered context, a soft ceiling on available context rather than on
response quality, kernel-managed injection statistically matches
performance on two of three models and shows a small, model-specific
deficit on the third, while using substantially shorter prompts and
correspondingly lower latency and token cost. These results support the
paper's main claim: centralizing memory retrieval, privacy enforcement,
and injection in the agent-system kernel delivers most of the
personalization benefit of unconstrained context at a fraction of its
cost, and does so more reliably than either an unmanaged memory backend
or standard retrieval augmentation.

\section{Related Work}

\paragraph{Memory systems for LLM agents.}
Prior work has largely focused on extending persistence beyond the
context window: MemoryBank~\citep{zhong2024memorybank}, MemGPT
\citep{packer2023memgpt}, LongMem~\citep{wang2023longmem}, and ReadAgent
\citep{lee2024readagent} each place memory control inside a single agent
or model wrapper. More recent systems improve memory's structure and
adaptability -- Mem0~\citep{chhikara2025mem0}, Zep~\citep{rasmussen2025zep},
LangMem~\citep{langmem2025}, and A-MEM~\citep{xu2025amem} -- but memory
remains an application-layer component in all of them; we instead treat
access, filtering, formatting, and injection as kernel-managed operations
shared across agents.

\paragraph{Multi-agent frameworks and shared state.}
Frameworks such as AutoGen~\citep{wu2023autogen}, CAMEL~\citep{li2023camel},
MetaGPT~\citep{hong2024metagpt}, and AgentVerse~\citep{chen2023agentverse}
focus on coordination and role specialization, representing shared state
through message passing or workflow-specific context, with
personalization typically reconstructed per agent. We instead introduce a
shared-memory abstraction letting multiple agents contribute to and
consume a unified user context under system-level policy control. Recent
work treats cross-agent memory sharing itself as a governance problem:
Collaborative Memory~\citep{rezazadeh2025collaborative} formalizes
access control via provenance-tagged bipartite graphs, SSGM
\citep{lam2026ssgm} proposes write-validation and read-filtering gates
against drift, and~\citet{yang2026attackerneededunintentionalcrossuser} show that cross-user leakage can
arise from benign interactions alone. Our visibility rule and
write-ordering barrier (Sections~\ref{sec:metadata-visibility},
\ref{sec:write-barrier}) target the same failure class through a simpler
mechanism -- static per-item metadata and a single sequence-numbered
barrier rather than bipartite permission graphs or drift-modeling gates.

\paragraph{Personalization and user modeling.}
Personalization in LLMs is commonly framed as retrieval, prompting,
fine-tuning, or user modeling~\citep{lewis2020retrieval,salemi2024lamp,
salemi2024optimizationmethodspersonalizinglarge,kumar2024longlamp,
zhang2024personalizedllm,liu2025surveypersonalizedlargelanguage,
tan2023usermodelingeralarge}, assuming a single application or inference path; none
addresses how a multi-agent system decides which facts are visible to
which agent, when to retrieve them, or how to inject them -- questions we
treat as kernel-level infrastructure.

\paragraph{OS-inspired agent infrastructure.}
AIOS~\citep{mei2024aios} separates agent applications from kernel-managed
services such as scheduling and memory management; OS-Copilot
\citep{wu2024oscopilot}, SWE-agent~\citep{yang2024sweagent}, and
OpenHands~\citep{wang2024openhands} similarly emphasize runtime
infrastructure over prompting alone, and MemOS~\citep{li2025memosmemoryosai} argues
memory should be schedulable. We build on this perspective but focus on a
narrower gap: existing infrastructure does not make shared personalization
memory the central kernel service through which agents obtain
user-specific state.

\section{Methodology and Architecture}

\subsection{Overview and System Model}
\label{sec:system-model}

We propose kernel-managed shared memory as a system-level abstraction for
personalization in multi-agent LLM systems. The central design choice is
to move personalization logic out of individual agents and into the
agent-system kernel: agents write structured memories with standardized
metadata, while the kernel manages identity resolution, write ordering,
retrieval, privacy filtering, ranking, formatting, token-budget control,
and prompt injection. Although our implementation is built on AIOS, the
abstraction is not specific to AIOS: any multi-agent runtime with a
shared memory backend and a controlled prompt-construction path can
implement the same design. We refer to the system layer generically as
the \emph{agent-system kernel}, with AIOS as our concrete instantiation.

We consider a multi-agent system with agents
$\mathcal{A} = \{a_1, a_2, \ldots, a_n\}$ and a memory store $\mathcal{M}$.
Each agent receives a query $q = (a_i, u, x)$, where $a_i$ is the
requesting agent, $u$ is the user identifier, and $x$ is the input
prompt. A memory item is $m = (u, o, t, s, c)$, where $o$ is the owner
agent, $t$ is the memory type, $s$ is the sharing policy, and $c$ is the
content. Memory types include \texttt{profile}, \texttt{task\_context},
and \texttt{conversation}; sharing policies are \texttt{private} or
\texttt{shared}.

\paragraph{Resolving user identity.}
Agents do not share a consistent notion of user identity independent of
the kernel, and naively using the requesting agent's own identifier as a
proxy produces cross-user contamination under concurrent trials. We
instead resolve $u$ through an explicit priority order -- an identifier
attached to the request, the most recently registered session
identifier, a fallback registry of previously observed identifiers, and
the requesting agent's own identifier as a last resort -- enforced
uniformly at the kernel boundary rather than left to per-agent
convention, closing a class of identity-resolution bugs encountered
during development (Section~\ref{sec:experiments}).

The kernel transforms the original prompt into an augmented prompt
$x' = K(a_i, u, x)$, where $K$ denotes the kernel-managed
personalization function.

\subsection{Kernel-Managed Shared Memory}
\label{sec:kernel-shared-memory}

\begin{figure}[htbp]
    \centering
    \includegraphics[width=0.62\linewidth]{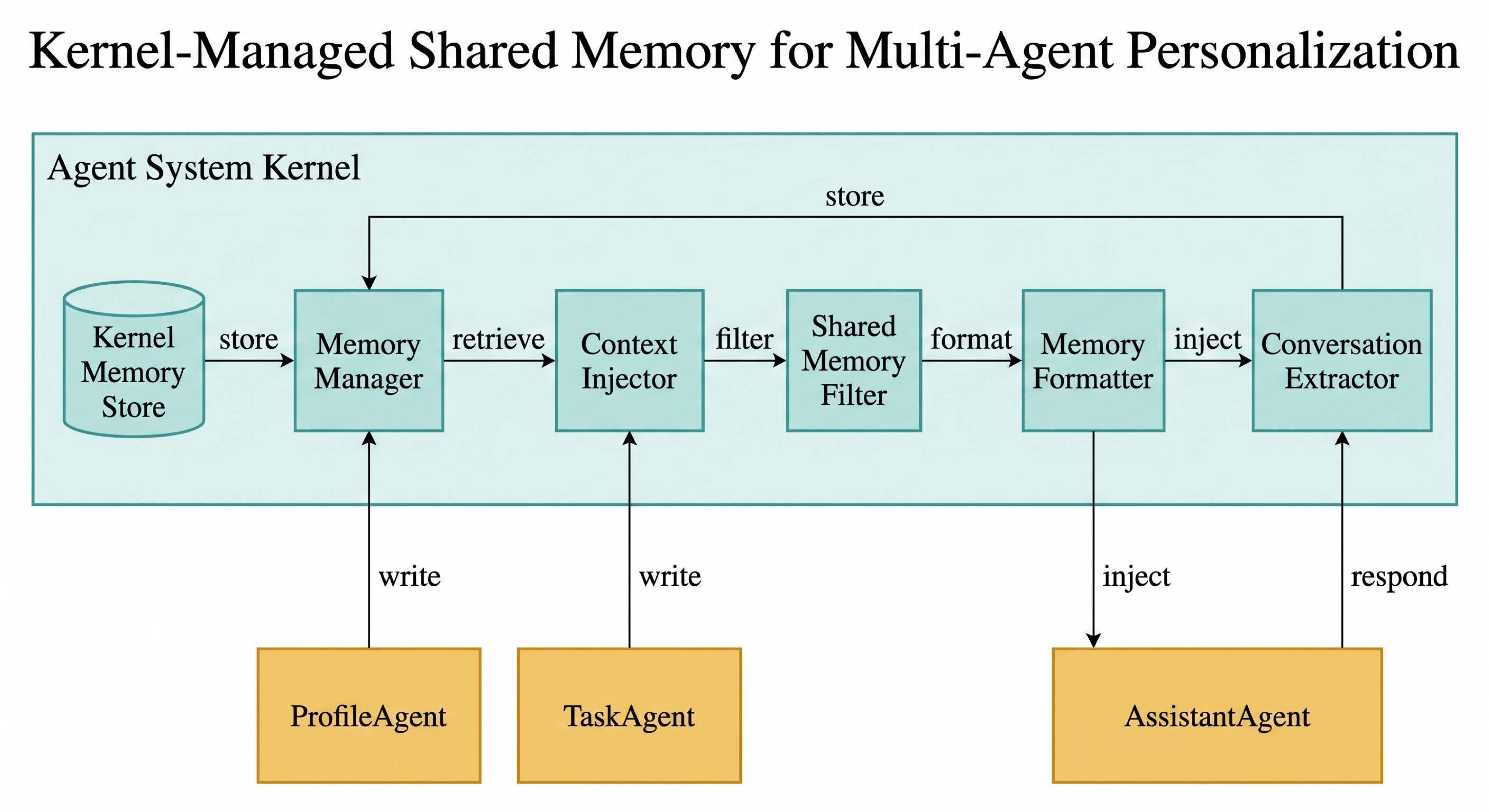}
    \caption{Kernel-managed shared memory: ProfileAgent and TaskAgent write
    structured memories, AssistantAgent remains retrieval-free, and the
    kernel enforces write ordering, privacy, retrieval, formatting, and
    injection between them.}
    \label{fig:kernel-architecture}
\end{figure}

Figure~\ref{fig:kernel-architecture} shows the proposed architecture.
ProfileAgent and TaskAgent act as memory-producing agents, extracting
stable user information and current task context, respectively.
AssistantAgent acts as a retrieval-free consumer: it does not query the
memory backend directly and does not construct its own personalization
prompt; all personalization context reaches it through the kernel-managed
prompt path.

The kernel implements personalization through five operations: (1)
enforce write-before-read ordering for the target user
(Section~\ref{sec:write-barrier}); (2) retrieve candidate memories,
$R = \mathrm{search}(\mathcal{M}, u, x)$; (3) filter retrieved memories
according to the system visibility rule
(Section~\ref{sec:metadata-visibility}); (4) rank and truncate the
filtered memories by semantic relevance and token budget; and (5) inject
the selected memories, $x' = \mathrm{inject}(x, R_k)$ -- part of the
system call path rather than application-specific prompt construction,
unlike agent-level memory systems.

\subsection{Memory Metadata and Visibility}
\label{sec:metadata-visibility}

Each memory item carries standardized metadata: \texttt{owner\_agent},
\texttt{user\_id}, \texttt{memory\_type}, and \texttt{sharing\_policy}.
The important design choice is that \texttt{owner\_agent} and
\texttt{user\_id} are independent: multiple agents may write memories
about the same user, but ownership and visibility remain explicit. The
kernel enforces memory visibility through the following rule:
\begin{equation}
    \mathrm{visible}(m, a_i) =
    \begin{cases}
        \mathrm{True} & \text{if } o = a_i, \\
        \mathrm{True} & \text{if } s = \text{``shared''}, \\
        \mathrm{False} & \text{otherwise.}
    \end{cases}
    \label{eq:visibility}
\end{equation}
If the sharing policy is absent, malformed, or explicitly private, the
memory is treated as private, making privacy a kernel invariant rather
than an SDK convention or prompt-level instruction.

\paragraph{Privacy invariant.}
Unlike the write-ordering guarantee of Section~\ref{sec:write-barrier},
which bounds a \emph{temporal} property, the visibility rule is a
\emph{static} property of memory metadata alone. Formally, for all
memories $m$ and agents $a_i \neq a_j$ with $o(m) = a_j$ and $s(m) =
\text{private}$,
\begin{equation}
    \mathrm{visible}(m, a_i) = \mathrm{False}
    \quad \text{for every execution history}.
    \label{eq:privacy-invariant}
\end{equation}
This is a safety property that holds unconditionally, unlike
Eq.~\ref{eq:write-barrier-guarantee}, which is a liveness/consistency
property holding only up to a bounded timeout.

\paragraph{Threat model and empirical verification.} Enforcing visibility
once, at the kernel, means an agent cannot exfiltrate memories merely by
omitting its own filtering logic, though this guarantee covers visibility
(not integrity) and is conditional on correct identity resolution
(full threat model in Appendix~\ref{sec:appendix-threat-model}). We
verify Eq.~\ref{eq:privacy-invariant} empirically at both configuration
endpoints: fully private ($n=450$) shows $0/450$ exposure, and fully
shared ($n=450$) shows no leakage of the private-by-default
\texttt{conversation} type among $1{,}798$ retrieved items.

\subsection{Cross-Agent Retrieval Pipeline and Write-Ordering Guarantee}
\label{sec:retrieval-pipeline}
\label{sec:write-barrier}

Given a user query, the kernel resolves the target \texttt{user\_id}
(Section~\ref{sec:system-model}), waits for any pending writes for that
user to be durably confirmed (below), retrieves candidate memories,
filters them by ownership and sharing policy, merges and deduplicates
results, ranks by semantic relevance, formats into natural language,
truncates to the token budget, and prepends the resulting memory block to
the assistant prompt. This exposes a unified user context to multiple
agents while preventing direct access to private agent-local memories, so
cross-agent personalization is achieved without requiring AssistantAgent
to implement retrieval, filtering, or prompt construction itself.

Because ProfileAgent and TaskAgent write memories asynchronously relative
to AssistantAgent's retrieval, a naive implementation is subject to a
race condition: retrieval may execute before writer agents' memories are
durably indexed, silently degrading personalization with no observable
error. We address this with a per-\texttt{user\_id} write barrier: each
write is stamped with a monotonically increasing sequence number at
acceptance time, scoped to the target \texttt{user\_id}; each retrieval
snapshots the highest sequence number issued for that user and blocks
until all writes up to that snapshot are confirmed drained, subject to a
bounded timeout (default 5000\,ms), after which it proceeds in a
fail-open mode that prioritizes liveness over strict consistency
(design rationale in Appendix~\ref{sec:appendix-barrier}).

Formally, let $\sigma(w)$ denote the sequence number assigned to write $w$
for user $u$, and let $\sigma^*(u)$ denote the highest sequence number
issued for $u$ at the time a retrieval $r$ begins. The barrier guarantees:
\begin{equation}
    r \text{ observes } \{w : o(w) = u,\ \sigma(w) \leq \sigma^*(u)\}
    \quad \text{or timeout}(u) \text{ elapses}.
    \label{eq:write-barrier-guarantee}
\end{equation}
This gives read-after-write consistency scoped to a single user, without
global locking or blocking retrievals for unrelated users. It eliminates
a class of nondeterministic injection failures, but does not by itself
guarantee correct scoping or formatting; we report its empirical
contribution, independent of those fixes, in
Appendix~\ref{sec:appendix-ablation}.

\subsection{Memory Representation, Formatting, and Agent Roles}
\label{sec:formatting}

Structured storage does not imply structured injection: in early
experiments, raw JSON memory content was difficult for smaller local
models to use reliably. The kernel therefore applies a formatting
function $c'_m = \mathrm{format}(c_m)$, rendering profile and task
memories as natural-language statements in place of raw JSON, while the
structured form is preserved for retrieval and auditing (isolated as an
independent ablation, Appendix~\ref{sec:appendix-ablation}).

We instantiate the method with three agents: ProfileAgent extracts stable
user attributes (preferences, tools, language, response style); TaskAgent
extracts short-term working context (goals, blockers, next steps); and
AssistantAgent generates user-facing responses from kernel-injected
context, performing no retrieval of its own. The method is implemented
as a kernel abstraction over LLM and memory operations, supporting
configurable extraction, injection, relevance thresholds, memory budgets,
write-barrier timeout, and pluggable memory providers, all inference-time
only with no retraining required.

\section{Experiments}
\label{sec:experiments}

\subsection{Baselines}
\label{sec:baselines}

A central question is whether observed gains come from the kernel's
specific architectural choices -- write ordering, identity resolution,
centralized privacy enforcement, cross-agent visibility -- or simply from
providing \emph{any} additional context. We compare against three
external baselines plus the proposed method (Table~\ref{tab:baselines}).
\texttt{naive\_concat} and \texttt{vanilla\_rag} involve no memory
system-level orchestration at all, and \texttt{mem0\_default} involves an
external memory system without kernel management, so any advantage of
\texttt{kernel\_shared} cannot be attributed merely to the presence of
injected context; comparing against \texttt{mem0\_default} specifically
isolates kernel-level write ordering, identity resolution, and privacy
enforcement, since both conditions use the same underlying provider.

\begin{table}[htbp]
    \centering
    \small
    \caption{Baselines compared against the proposed method.}
    \label{tab:baselines}
    \begin{tabular}{lp{10.2cm}}
        \toprule
        Method & Description \\
        \midrule
        \texttt{naive\_concat} & Full synthetic profile/task context concatenated into the prompt; no retrieval or filtering. An upper bound on \emph{available} context, not response quality. \\
        \texttt{vanilla\_rag} & Profile/task context chunked and indexed; top-$k$ chunks retrieved via embedding search. Standard RAG absent a dedicated memory system. \\
        \texttt{mem0\_default} & Memory via Mem0, configured private-only, with no kernel orchestration, write ordering, or cross-agent sharing. \\
        \texttt{kernel\_shared} & Proposed method: ProfileAgent/TaskAgent write \texttt{shared} memories; kernel enforces ordering, identity resolution, visibility, and cross-agent injection. \\
        \bottomrule
    \end{tabular}
\end{table}

\subsection{Models, Trials, Evaluation, and Configuration}
\label{sec:evaluation}

We evaluate all four methods across three assistant models of varying
capability -- GPT-4o, Llama-3.1:8B, and Qwen-2.5:7B -- to distinguish
architectural effects from effects better explained by capability alone.
Each trial generates a synthetic user profile, a synthetic task context,
and a vague follow-up query (e.g., ``What should I focus on next?''),
intentionally underspecified so personalized content must come from
injected memory rather than the query itself.

Each trial is scored by GPT-5.4 along three dimensions on a 1--5 scale:
\textbf{Profile Usage}, \textbf{Task Usage}, and \textbf{Integration}
(whether profile and task information are combined coherently). We use
GPT-5.4, a model distinct from every assistant evaluated, as the sole
judge across all twelve (model, method) conditions to avoid a
self-preference effect in LLM-as-judge evaluation.

The kernel ran with \texttt{auto\_extract: true}, \texttt{auto\_inject:
true}, a relevance threshold of $0.3$, a maximum of $10$ injected
memories per call, a memory token budget of $2000$, and the write barrier
(Section~\ref{sec:write-barrier}) enabled with a $5000$\,ms timeout. Each
of the twelve conditions was run for $150$ trials ($1{,}800$ total).

\subsection{Human Validation}
\label{sec:human-validation}

To validate the automated judge, we constructed a blinded human-rated
subset (30 GPT-4o trials, 24 unique items plus 6 duplicates for
intra-rater consistency; design in
Appendix~\ref{sec:appendix-human-validation}). Human and automated judge
scores matched exactly on $41.7\%$ of items and agreed within one point
on $83.3\%$ (mean absolute error $0.75$), and were strongly correlated
(Spearman's $\rho = 0.727$, $p = 5.7 \times 10^{-5}$). The human rater
confirms our central GPT-4o comparison: \texttt{kernel\_shared} and
\texttt{naive\_concat} are indistinguishable in human-perceived quality
($4.17$ vs.\ $4.17$), both far above \texttt{mem0\_default} ($1.00$) --
an independent confirmation not filtered through the automated judge.
The judge does under-credit \texttt{vanilla\_rag} relative to the human
rater ($2.17$ vs.\ $3.67$), consistent with the architectural failure
mode in Section~\ref{sec:results} where \texttt{vanilla\_rag}
systematically drops profile content; per-method agreement and
intra-rater consistency detail are in
Appendix~\ref{sec:appendix-human-validation}.

\subsection{Quantitative Results}
\label{sec:results}

Table~\ref{tab:main-results} reports mean scores with standard deviations
across all twelve (model, method) conditions, 150 trials each.

\begin{table}[htbp]
    \centering
    \scriptsize
    \setlength{\tabcolsep}{4pt}
    \renewcommand{\arraystretch}{0.65}
    \caption{Full results across 3 models $\times$ 4 methods, 150 trials per condition, judged by GPT-5.4. Values are mean $\pm$ SD.}
    \label{tab:main-results}
    \begin{tabular}{llccccc}
        \toprule
        Model & Method & Profile & Task & Integration & Latency (s) \\
        \midrule
            GPT-4o & \texttt{naive\_concat}  & $4.61 \pm 0.67$ & $4.93 \pm 0.25$ & $4.50 \pm 0.63$ & 19.2 \\
        GPT-4o & \texttt{kernel\_shared}  & $4.69 \pm 0.50$ & $4.99 \pm 0.08$ & $4.61 \pm 0.52$ & 7.5 \\
        GPT-4o & \texttt{vanilla\_rag}    & $1.84 \pm 0.56$ & $3.84 \pm 0.54$ & $1.96 \pm 0.40$ & 10.8 \\
        GPT-4o & \texttt{mem0\_default}   & $1.05 \pm 0.21$ & $1.03 \pm 0.16$ & $1.00 \pm 0.00$ & 3.9 \\[2pt]
        Llama-3.1:8B & \texttt{naive\_concat} & $4.03 \pm 0.74$ & $4.54 \pm 0.50$ & $3.93 \pm 0.65$ & 25.5 \\
        Llama-3.1:8B & \texttt{kernel\_shared} & $3.87 \pm 0.83$ & $4.33 \pm 0.75$ & $3.63 \pm 0.80$ & 21.7 \\
        Llama-3.1:8B & \texttt{vanilla\_rag}   & $1.91 \pm 0.74$ & $3.78 \pm 0.62$ & $1.99 \pm 0.72$ & 18.9 \\
        Llama-3.1:8B & \texttt{mem0\_default}  & $1.07 \pm 0.25$ & $1.31 \pm 0.47$ & $1.04 \pm 0.20$ & 10.5 \\[2pt]
        Qwen-2.5:7B & \texttt{naive\_concat} & $3.71 \pm 1.17$ & $4.43 \pm 0.54$ & $3.61 \pm 0.94$ & 36.3 \\
        Qwen-2.5:7B & \texttt{kernel\_shared} & $3.63 \pm 1.09$ & $4.40 \pm 0.49$ & $3.58 \pm 0.94$ & 26.3 \\
        Qwen-2.5:7B & \texttt{vanilla\_rag}   & $1.84 \pm 0.72$ & $3.84 \pm 0.51$ & $1.89 \pm 0.56$ & 32.8 \\
        Qwen-2.5:7B & \texttt{mem0\_default}  & $1.24 \pm 0.43$ & $1.60 \pm 0.51$ & $1.19 \pm 0.39$ & 26.9 \\
        \bottomrule
    \end{tabular}
    \renewcommand{\arraystretch}{0.85}
\end{table}

Three patterns hold across all three models. First,
\texttt{vanilla\_rag} profile scores stay flat regardless of model
capability ($1.84$/$1.91$/$1.84$) while task scores are both higher and
similarly flat ($3.84$/$3.78$/$3.84$) -- an architectural failure, not a
capability one: the follow-up query is closer to task content, so
retrieval consistently drops profile chunks regardless of the model's
capacity to use them if present. Second, \texttt{mem0\_default} scores
near floor throughout; since it shares \texttt{kernel\_shared}'s
underlying provider, this isolates write-ordering and identity-resolution
guarantees, not the mere presence of a memory backend, as the source of
the gap. Third, \texttt{kernel\_shared} vs.\ \texttt{naive\_concat} is
model-dependent (Welch's $t$, Mann--Whitney $U$, bootstrap CI; full nine
model$\times$dimension comparisons and figure in
Appendix~\ref{sec:appendix-stats}): it ties \texttt{naive\_concat} on
GPT-4o Profile/Integration and all three Qwen-2.5:7B dimensions, is
higher on GPT-4o Task ($p=0.006$), and is significantly \emph{lower} on
Llama-3.1:8B Task/Integration (rank-biserial $0.13$--$0.19$, unexplained
by scale).
Against \texttt{vanilla\_rag} and \texttt{mem0\_default}, by contrast,
the advantage is large and overwhelmingly significant everywhere (all
$p<10^{-18}$, rank-biserial $0.44$--$1.00$): kernel-managed injection's
downside is capped at a small, model-specific gap against unfiltered
context, while its advantage over an unmanaged backend or plain retrieval
is categorical.

\subsection{Latency Analysis}
\label{sec:latency}

\texttt{kernel\_shared} achieves substantially lower end-to-end latency
than \texttt{naive\_concat} on all three models: $7.5$s vs.\ $19.2$s on
GPT-4o, $21.7$s vs.\ $25.5$s on Llama-3.1:8B, and $26.3$s vs.\ $36.3$s on
Qwen-2.5:7B ($15$--$61\%$ reduction). The mechanism differs by model
(full analysis in Appendix~\ref{sec:appendix-latency}): on Llama-3.1:8B,
output length does not differ significantly between conditions
($p=0.27$) yet \texttt{kernel\_shared} is still $15\%$ faster, pointing
to input-side savings; on GPT-4o and Qwen-2.5:7B, \texttt{naive\_concat}
produces significantly longer outputs ($+159$ and $+83$ words,
$p<0.0001$), accounting for part of the gap. \texttt{mem0\_default}'s low
latency simply reflects little content to inject. This latency win holds
regardless of outcome on the \texttt{naive\_concat} quality comparison
above.

\section{Discussion and Conclusion}

Kernel-managed shared memory reliably outperforms a standard RAG pipeline
and an unmanaged external memory backend across all three models, at
lower latency than full unfiltered context; the \texttt{naive\_concat}
comparison is more nuanced, tying on two of three models with a
Llama-3.1:8B deficit unexplained by scale, left to future work.

\begin{ack}
We thank the AIOS project contributors and collaborators whose prior architecture and tooling made this extension possible.
\end{ack}

\newpage

\bibliographystyle{plainnat}
\bibliography{references}

@article{mei2024aios,
  title={AIOS: LLM Agent Operating System},
  author={Mei, Kai and Zhu, Xi and Xu, Wujiang and Jin, Mingyu and Hua, Wenyue and Li, Zelong and Xu, Shuyuan and Ye, Ruosong and Ge, Yingqiang and Zhang, Yongfeng},
  journal={arXiv preprint arXiv:2403.16971},
  year={2024},
  url={https://arxiv.org/abs/2403.16971}
}

@article{packer2023memgpt,
  title={MemGPT: Towards LLMs as Operating Systems},
  author={Packer, Charles and Wooders, Sarah and Lin, Kevin and Fang, Vivian and Patil, Shishir G. and Stoica, Ion and Gonzalez, Joseph E.},
  journal={arXiv preprint arXiv:2310.08560},
  year={2023},
  url={https://arxiv.org/abs/2310.08560}
}

@article{xu2025amem,
  title={A-MEM: Agentic Memory for LLM Agents},
  author={Xu, Wujiang and Liang, Zujie and Mei, Kai and Gao, Hang and Tan, Juntao and Zhang, Yongfeng},
  journal={arXiv preprint arXiv:2502.12110},
  year={2025},
  url={https://arxiv.org/abs/2502.12110}
}

@article{chhikara2025mem0,
  title={Mem0: Building Production-Ready AI Agents with Scalable Long-Term Memory},
  author={Chhikara, Prateek and Khant, Dev and Aryan, Saket and Singh, Taranjeet and Yadav, Deshraj},
  journal={arXiv preprint arXiv:2504.19413},
  year={2025},
  url={https://arxiv.org/abs/2504.19413}
}

@inproceedings{zhong2024memorybank,
  title={MemoryBank: Enhancing Large Language Models with Long-Term Memory},
  author={Zhong, Wanjun and Guo, Lianghong and Gao, Qiqi and Ye, He and Wang, Yanlin},
  booktitle={Proceedings of the AAAI Conference on Artificial Intelligence},
  volume={38},
  number={17},
  pages={19724--19731},
  year={2024},
  doi={10.1609/aaai.v38i17.29946}
}

@article{wang2023longmem,
  title={Augmenting Language Models with Long-Term Memory},
  author={Wang, Weizhi and Dong, Li and Cheng, Hao and Liu, Xiaodong and Yan, Xifeng and Gao, Jianfeng and Wei, Furu},
  journal={arXiv preprint arXiv:2306.07174},
  year={2023},
  url={https://arxiv.org/abs/2306.07174}
}

@article{lee2024readagent,
  title={A Human-Inspired Reading Agent with Gist Memory of Very Long Contexts},
  author={Lee, Kuang-Huei and Chen, Xinyun and Furuta, Hiroki and Canny, John and Fischer, Ian},
  journal={arXiv preprint arXiv:2402.09727},
  year={2024},
  url={https://arxiv.org/abs/2402.09727}
}

@misc{langmem2025,
  title={LangMem: Long-Term Memory for Agents},
  author={{LangChain Team}},
  year={2025},
  howpublished={\url{https://langchain-ai.github.io/langmem/}},
  note={Accessed 2026-04-25}
}

@article{rasmussen2025zep,
  title={Zep: A Temporal Knowledge Graph Architecture for Agent Memory},
  author={Rasmussen, Preston and Paliychuk, Pavlo and Beauvais, Travis and Ryan, Jack and Chalef, Daniel},
  journal={arXiv preprint arXiv:2501.13956},
  year={2025},
  url={https://arxiv.org/abs/2501.13956}
}

@article{wu2023autogen,
  title={AutoGen: Enabling Next-Gen LLM Applications via Multi-Agent Conversation Framework},
  author={Wu, Qingyun and Bansal, Gagan and Zhang, Jieyu and Wu, Yiran and Li, Beibin and Zhu, Erkang and Jiang, Li and Zhang, Xiaoyun and Zhang, Shaokun and Liu, Jiale and Awadallah, Ahmed Hassan and White, Ryen W. and Burger, Doug and Wang, Chi},
  journal={arXiv preprint arXiv:2308.08155},
  year={2023},
  url={https://arxiv.org/abs/2308.08155}
}

@inproceedings{li2023camel,
  title={CAMEL: Communicative Agents for ``Mind'' Exploration of Large Language Model Society},
  author={Li, Guohao and Hammoud, Hasan Abed Al Kader and Itani, Hani and Khizbullin, Dmitrii and Ghanem, Bernard},
  booktitle={Advances in Neural Information Processing Systems},
  year={2023},
  url={https://arxiv.org/abs/2303.17760}
}

@inproceedings{hong2024metagpt,
  title={MetaGPT: Meta Programming for A Multi-Agent Collaborative Framework},
  author={Hong, Sirui and Zhuge, Mingchen and Chen, Jonathan and Zheng, Xiawu and Cheng, Yuheng and Wang, Jinlin and Zhang, Ceyao and Wang, Zili and Yau, Steven Ka Shing and Lin, Zijuan and others},
  booktitle={International Conference on Learning Representations},
  year={2024},
  url={https://arxiv.org/abs/2308.00352}
}

@article{chen2023agentverse,
  title={AgentVerse: Facilitating Multi-Agent Collaboration and Exploring Emergent Behaviors},
  author={Chen, Weize and Su, Yusheng and Zuo, Jingwei and Yang, Cheng and Yuan, Chenfei and Chan, Chi-Min and Yu, Heyang and Lu, Yaxi and Hung, Yi-Hsin and Qian, Chen and Qin, Yujia and Cong, Xin and Xie, Ruobing and Liu, Zhiyuan and Sun, Maosong and Zhou, Jie},
  journal={arXiv preprint arXiv:2308.10848},
  year={2023},
  url={https://arxiv.org/abs/2308.10848}
}

@inproceedings{lewis2020retrieval,
  title={Retrieval-Augmented Generation for Knowledge-Intensive NLP Tasks},
  author={Lewis, Patrick and Perez, Ethan and Piktus, Aleksandra and Petroni, Fabio and Karpukhin, Vladimir and Goyal, Naman and K{\"u}ttler, Heinrich and Lewis, Mike and Yih, Wen-tau and Rockt{\"a}schel, Tim and Riedel, Sebastian and Kiela, Douwe},
  booktitle={Advances in Neural Information Processing Systems},
  volume={33},
  pages={9459--9474},
  year={2020}
}

@inproceedings{salemi2024lamp,
  title={LaMP: When Large Language Models Meet Personalization},
  author={Salemi, Alireza and Mysore, Sheshera and Bendersky, Michael and Zamani, Hamed},
  booktitle={Proceedings of the 62nd Annual Meeting of the Association for Computational Linguistics},
  year={2024},
  url={https://arxiv.org/abs/2304.11406}
}

@article{kumar2024longlamp,
  title={LongLaMP: A Benchmark for Personalized Long-Form Text Generation},
  author={Kumar, Ishita and Viswanathan, Snigdha and Yerra, Sushrita and Salemi, Alireza and Rossi, Ryan A. and Dernoncourt, Franck and Deilamsalehy, Hanieh and Chen, Xiang and Zhang, Ruiyi and Agarwal, Shubham and others},
  journal={arXiv preprint arXiv:2407.11016},
  year={2024},
  url={https://arxiv.org/abs/2407.11016}
}

@article{zhang2024personalizedllm,
  title={Personalization of Large Language Models: A Survey},
  author={Zhang, Zhehao and Rossi, Ryan A. and Kveton, Branislav and Shao, Yijia and Yang, Diyi and Zamani, Hamed and Dernoncourt, Franck and Barrow, Joe and Yu, Tong and Kim, Sungchul and others},
  journal={arXiv preprint arXiv:2411.00027},
  year={2024},
  url={https://arxiv.org/abs/2411.00027}
}

@misc{tan2023usermodelingeralarge,
      title={User Modeling in the Era of Large Language Models: Current Research and Future Directions}, 
      author={Zhaoxuan Tan and Meng Jiang},
      year={2023},
      eprint={2312.11518},
      archivePrefix={arXiv},
      primaryClass={cs.CL},
      url={https://arxiv.org/abs/2312.11518}, 
}

@article{wu2024oscopilot,
  title={OS-Copilot: Towards Generalist Computer Agents with Self-Improvement},
  author={Wu, Zhiyong and Han, Chengcheng and Ding, Zichen and Weng, Zhenmin and Liu, Zhoumianze and Yao, Shunyu and Yu, Tao and Kong, Lingpeng},
  journal={arXiv preprint arXiv:2402.07456},
  year={2024},
  url={https://arxiv.org/abs/2402.07456}
}

@inproceedings{yang2024sweagent,
  title={SWE-agent: Agent-Computer Interfaces Enable Automated Software Engineering},
  author={Yang, John and Jimenez, Carlos E. and Wettig, Alexander and Lieret, Kilian and Yao, Shunyu and Narasimhan, Karthik and Press, Ofir},
  booktitle={Advances in Neural Information Processing Systems},
  year={2024},
  url={https://arxiv.org/abs/2405.15793}
}

@article{wang2024openhands,
  title={OpenHands: An Open Platform for AI Software Developers as Generalist Agents},
  author={Wang, Xingyao and Li, Boxuan and Song, Yufan and Xu, Frank F. and Tang, Xiangru and Zhuge, Mingchen and Pan, Jiayi and Song, Yueqi and Li, Bowen and Singh, Jaskirat and others},
  journal={arXiv preprint arXiv:2407.16741},
  year={2024},
  url={https://arxiv.org/abs/2407.16741}
}

@misc{li2025memosmemoryosai,
      title={MemOS: A Memory OS for AI System}, 
      author={Zhiyu Li and Chenyang Xi and Chunyu Li and Ding Chen and Boyu Chen and Shichao Song and Simin Niu and Hanyu Wang and Jiawei Yang and Chen Tang and Qingchen Yu and Jihao Zhao and Yezhaohui Wang and Peng Liu and Zehao Lin and Pengyuan Wang and Jiahao Huo and Tianyi Chen and Kai Chen and Kehang Li and Zhen Tao and Huayi Lai and Hao Wu and Bo Tang and Zhengren Wang and Zhaoxin Fan and Ningyu Zhang and Linfeng Zhang and Junchi Yan and Mingchuan Yang and Tong Xu and Wei Xu and Huajun Chen and Haofen Wang and Hongkang Yang and Wentao Zhang and Zhi-Qin John Xu and Siheng Chen and Feiyu Xiong},
      year={2025},
      eprint={2507.03724},
}

@article{rezazadeh2025collaborative,
  title={Collaborative Memory: Multi-User Memory Sharing in LLM Agents with Dynamic Access Control},
  author={Rezazadeh, Alireza and Li, Zichao and Lou, Ange and Zhao, Yuying and Wei, Wei and Bao, Yujia},
  journal={arXiv preprint arXiv:2505.18279},
  year={2025}
}

@article{lam2026ssgm,
  title={Governing Evolving Memory in {LLM} Agents: Risks, Mechanisms, and the Stability and Safety Governed Memory ({SSGM}) Framework},
  author={Lam, Chingkwun and Li, Jiaxin and Zhang, Lingfei and Zhao, Kuo},
  journal={arXiv preprint arXiv:2603.11768},
  year={2026}
}

@misc{yang2026attackerneededunintentionalcrossuser,
      title={No Attacker Needed: Unintentional Cross-User Contamination in Shared-State LLM Agents}, 
      author={Tiankai Yang and Jiate Li and Yi Nian and Shen Dong and Ruiyao Xu and Ryan Rossi and Kaize Ding and Yue Zhao},
      year={2026},
      eprint={2604.01350},
      archivePrefix={arXiv},
      primaryClass={cs.CL},
      url={https://arxiv.org/abs/2604.01350}, 
}

@misc{salemi2024optimizationmethodspersonalizinglarge,
      title={Optimization Methods for Personalizing Large Language Models through Retrieval Augmentation}, 
      author={Alireza Salemi and Surya Kallumadi and Hamed Zamani},
      year={2024},
      eprint={2404.05970},
      archivePrefix={arXiv},
      primaryClass={cs.CL},
      url={https://arxiv.org/abs/2404.05970}, 
}

@misc{liu2025surveypersonalizedlargelanguage,
      title={A Survey of Personalized Large Language Models: Progress and Future Directions}, 
      author={Jiahong Liu and Zexuan Qiu and Zhongyang Li and Quanyu Dai and Wenhao Yu and Jieming Zhu and Minda Hu and Menglin Yang and Tat-Seng Chua and Irwin King},
      year={2025},
      eprint={2502.11528},
      archivePrefix={arXiv},
      primaryClass={cs.AI},
      url={https://arxiv.org/abs/2502.11528}, 
}

\appendix

\section{Technical Appendix}

\subsection{Human Validation Design Detail}
\label{sec:appendix-human-validation}

We sampled 30 GPT-4o trials: 6 unique items per method across all four
methods (24 unique items), plus 6 duplicate items inserted at separated
positions to measure intra-rater consistency. Items were presented in
randomized order under opaque identifiers, with the rater blind to both
condition and automated judge score. A single rater scored each item on
the Integration dimension using the same 1--5 rubric given to the
automated judge.

\begin{table}[htbp]
    \centering
    \footnotesize
    \caption{Human rater vs.\ automated judge agreement, by method (GPT-4o, Integration dimension, $N=6$ per method).}
    \label{tab:human-validation}
    \begin{tabular}{lccc}
        \toprule
        Method & Human mean & Judge mean & Bias (H$-$J) \\
        \midrule
        \texttt{mem0\_default}  & 1.00 & 1.00 & $0.00$  \\
        \texttt{kernel\_shared} & 4.17 & 4.50 & $-0.33$ \\
        \texttt{naive\_concat}  & 4.17 & 4.67 & $-0.50$ \\
        \texttt{vanilla\_rag}   & 3.67 & 2.17 & $+1.50$ \\
        \bottomrule
    \end{tabular}
\end{table}

Intra-rater consistency on the 6 duplicate pairs was high: the rater gave
the exact same score on 5 of 6 repeats, with a mean absolute difference of
$0.17$ across all six pairs. We note the use of a single rater precludes
computing inter-rater reliability statistics such as Krippendorff's
$\alpha$. Read together, the judge appears to be a trustworthy proxy for
the paper's central claim, since its one documented miscalibration
under-credits a baseline rather than over-crediting the proposed method.

\subsection{Threat Model and Empirical Verification Detail}
\label{sec:appendix-threat-model}

\paragraph{Threat model.} We consider an adversary model in which one
agent -- compromised, buggy, or simply malicious -- attempts to read
memories it is not authorized to see: either another agent's private
memories for the same user, or any agent's memories for a different user
entirely. Because visibility is checked once, at the kernel, on every
retrieval path (Eq.~\ref{eq:privacy-invariant}), such an agent cannot
exfiltrate private memories simply by skipping its own filtering logic;
it would need to compromise the kernel's enforcement point directly, a
materially higher bar than omitting a check in one of $n$
independently-trusted implementations. This protection has two concrete
limits. First, the kernel enforces \emph{visibility}, not
\emph{integrity} or \emph{provenance}: a malicious ProfileAgent can still
corrupt what AssistantAgent believes about a user. Second, the kernel
enforces visibility \emph{given a resolved user identity}; it does not
independently verify that the identity presented to it is correct. If the
identity-resolution fallback chain (Section~\ref{sec:system-model}) can
be induced to resolve to the wrong \texttt{user\_id} -- for instance, by
an agent that races to register itself as the most recent session for a
target user before the legitimate session does -- the visibility rule
would faithfully enforce access control for the \emph{wrong} partition, a
failure of identity resolution rather than of visibility enforcement, but
with the same practical consequence. We have not evaluated the
identity-resolution fallback chain against an adversarial agent
attempting this.

\paragraph{Empirical verification at both configuration endpoints.} We
have not run a designed sweep over intermediate sharing configurations or
an adversarial query campaign; both are future work. Our existing
benchmark trials (Section~\ref{sec:experiments}) provide a zero-cost
empirical check of Eq.~\ref{eq:privacy-invariant} at the two endpoints of
the sharing policy. At the fully-private endpoint
(\texttt{mem0\_default}, $n=450$ trials across three models, all memories
written \texttt{private}), retrieval-eligible cross-agent exposure is
$0/450$. Functional exposure is $0/450$ as well, with one apparent
exception we verified by hand: a single trial scored $3/5$ on Task Usage
despite zero memories retrieved, which we trace to the automated judge
crediting generic clarifying language rather than any genuinely leaked
content -- a judge-calibration artifact, not a privacy failure. At the
fully-shared endpoint (\texttt{kernel\_shared}, $n=450$ trials), we
additionally checked whether sharing one memory type inadvertently widens
exposure of another: of $1{,}798$ total memory items retrieved across all
trials, none were the private-by-default \texttt{conversation} type
written by the \texttt{ConversationExtractor}
(Appendix~\ref{sec:appendix-syscalls}), confirming that
\texttt{sharing\_policy} scoping is enforced per memory item rather than
per writer agent.

\subsection{Significance Testing Detail}
\label{sec:appendix-stats}

\begin{figure}[htbp]
    \centering
    \includegraphics[width=0.42\linewidth]{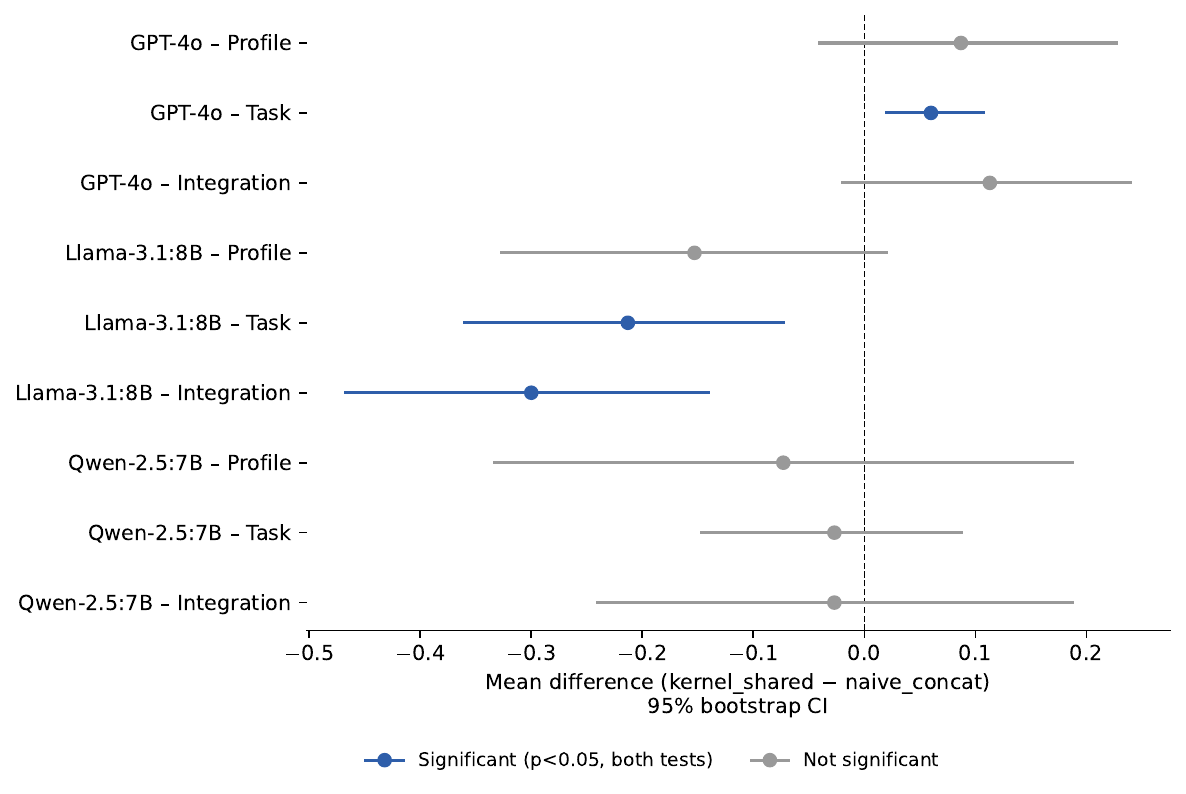}
    \caption{Mean difference and 95\% bootstrap CI for the nine
    \texttt{kernel\_shared} vs.\ \texttt{naive\_concat} comparisons; six of
    nine intervals cross zero (dashed line), indicating no significant
    difference.}
    \label{fig:significance-forest}
\end{figure}

Table~\ref{tab:significance} and Figure~\ref{fig:significance-forest}
report the full nine (model, dimension) \texttt{kernel\_shared} vs.\
\texttt{naive\_concat} comparisons underlying the discussion in
Section~\ref{sec:results}.

\begin{table}[h]
    \centering
    \small
    \caption{\texttt{kernel\_shared} vs.\ \texttt{naive\_concat}: mean
    difference and significance across all three models (independent
    samples, $n=150$ per condition). Bold marks $p < 0.05$ under both
    Welch's $t$-test and Mann--Whitney $U$.}
    \label{tab:significance}
    \begin{tabular}{llrrrr}
        \toprule
        Model & Dimension & Mean $\Delta$ & Welch $p$ & MWU $p$ & Bootstrap 95\% CI \\
        \midrule
        GPT-4o & Profile & $+0.087$ & $0.209$ & $0.372$ & $[-0.040, +0.227]$ \\
        GPT-4o & Task & $\mathbf{+0.060}$ & $\mathbf{0.006}$ & $\mathbf{0.006}$ & $[+0.020, +0.107]$ \\
        GPT-4o & Integration & $+0.113$ & $0.090$ & $0.163$ & $[-0.020, +0.240]$ \\
        \addlinespace
        Llama-3.1:8B & Profile & $-0.153$ & $0.093$ & $0.088$ & $[-0.327, +0.020]$ \\
        Llama-3.1:8B & Task & $\mathbf{-0.213}$ & $\mathbf{0.004}$ & $\mathbf{0.026}$ & $[-0.360, -0.073]$ \\
        Llama-3.1:8B & Integration & $\mathbf{-0.300}$ & $\mathbf{0.0004}$ & $\mathbf{0.0001}$ & $[-0.467, -0.140]$ \\
        \addlinespace
        Qwen-2.5:7B & Profile & $-0.073$ & $0.575$ & $0.361$ & $[-0.333, +0.187]$ \\
        Qwen-2.5:7B & Task & $-0.027$ & $0.653$ & $0.530$ & $[-0.147, +0.087]$ \\
        Qwen-2.5:7B & Integration & $-0.027$ & $0.806$ & $0.755$ & $[-0.240, +0.187]$ \\
        \bottomrule
    \end{tabular}
\end{table}

\subsection{Latency Mechanism Detail}
\label{sec:appendix-latency}

Table~\ref{tab:output-length} reports output length (word count) by
condition and model, alongside a Welch's $t$-test on the
\texttt{kernel\_shared} vs.\ \texttt{naive\_concat} difference and the
within-condition correlation between output length and latency.

\begin{table}[h]
    \centering
    \small
    \caption{Output length (words) by condition and model, with
    significance of the \texttt{kernel\_shared} vs.\ \texttt{naive\_concat}
    difference and the within-condition correlation between output length
    and latency.}
    \label{tab:output-length}
    \begin{tabular}{lrrrrr}
        \toprule
        Model & \texttt{kernel\_shared} & \texttt{naive\_concat} & Diff & Welch $p$ & $r$(words, latency) \\
        \midrule
        GPT-4o        & $404.7$ & $563.8$ & $\mathbf{-159.1}$ & $\mathbf{<10^{-6}}$ & $0.66$ / $0.59$ \\
        Llama-3.1:8B  & $202.4$ & $193.8$ & $+8.6$            & $0.271$             & $0.84$ / $0.52$ \\
        Qwen-2.5:7B   & $218.7$ & $301.8$ & $\mathbf{-83.0}$  & $\mathbf{<10^{-4}}$ & $0.86$ / $0.83$ \\
        \bottomrule
    \end{tabular}
    \vspace{2pt}
    \\ \footnotesize $r$(words, latency) reported as \texttt{kernel\_shared} / \texttt{naive\_concat}, both $p < 10^{-11}$.
\end{table}

Within every condition, word count and latency are strongly correlated
($r=0.52$--$0.86$, all $p<10^{-11}$), confirming generation length is a
major latency driver in general; the relevant question is whether that
driver differs \emph{systematically between conditions}, which it does
for GPT-4o and Qwen-2.5:7B but not for Llama-3.1:8B (no significant
difference, $p=0.27$, yet \texttt{kernel\_shared} remains $15\%$ faster --
implicating input-side savings rather than generation length for that
model). \texttt{mem0\_default}'s short, generic outputs follow the same
pattern of low content, low latency, and are excluded from the table
since they are not a meaningful comparison point. One plausible reading
of the GPT-4o and Qwen-2.5:7B asymmetry is that unfiltered, unstructured
context prompts the model toward longer, more exhaustive responses, while
curated, kernel-formatted context elicits a more targeted one; we did not
design an experiment specifically to test this explanation and note it
as a plausible mechanism rather than a demonstrated one, for future work.

\subsection{Ablation Pilot Detail}
\label{sec:appendix-ablation}

Table~\ref{tab:ablation} reports the full pilot ablation deltas
underlying the pilot ablation study conducted during development. Raw
JSON injection, evaluated with an LLM-only judge, scored worse than the
private baseline on all three dimensions (row 1); natural-language
formatting and an explicit system-prompt instruction each narrowed this
gap without closing it (rows 2--3); only adopting the hybrid keyword+LLM
judge produced a positive result (row 4). This pilot (Qwen-2.5:7B, 30
trials) predates the verification pass applied to our main results and
carries no evidentiary weight toward the paper's central claims -- it is
included for its qualitative documentation of the design trajectory only.

\begin{table}[h]
    \centering
    \caption{Ablation: Phase 2 (shared) $-$ Phase 1 (private) score deltas
    across successive pipeline and evaluation refinements (Qwen-2.5:7B, 30-trial pilot).}
    \label{tab:ablation}
    \begin{tabular}{lccc}
        \toprule
        Configuration & Profile $\Delta$ & Task $\Delta$ & Integration $\Delta$ \\
        \midrule
        Raw JSON injection, LLM-only judge          & $-0.73$ & $-1.47$ & $-0.77$ \\
        + Natural-language formatting                & $-0.10$ & $-0.57$ & $-0.43$ \\
        + Explicit system-prompt instruction          & $-0.10$ & $-0.23$ & $-0.10$ \\
        + HybridJudge (30 trials)                     & $+1.37$ & $+1.27$ & $+0.87$ \\
        \bottomrule
    \end{tabular}
\end{table}

\subsection{System Call Interface}
\label{sec:appendix-syscalls}

Table~\ref{tab:appendix-syscalls} lists the kernel modules involved in
personalization and their corresponding operations, in the style of the
system-call catalog used by prior OS-inspired agent infrastructure
(e.g., AIOS's kernel syscall table).

\begin{table}[h]
    \centering
    \small
    \caption{Kernel modules and their corresponding operations for shared-memory personalization.}
    \label{tab:appendix-syscalls}
    \begin{tabular}{ll}
        \toprule
        Module & Operations \\
        \midrule
        MemoryManager & \texttt{address\_request}, \texttt{register\_user\_id}, \texttt{stamp\_barrier} \\
        Mem0Provider & \texttt{add\_memory}, \texttt{retrieve\_memory}, \texttt{get\_all}, \texttt{apply\_sharing\_filter} \\
        ContextInjector & \texttt{inject}, \texttt{resolve\_user\_id}, \texttt{merge\_and\_dedupe} \\
        MemoryWriteBarrier & \texttt{acquire}, \texttt{release}, \texttt{snapshot}, \texttt{wait\_until\_drained} \\
        MemoryFormatter & \texttt{format\_memory} \\
        ConversationExtractor & \texttt{extract\_async} \\
        \bottomrule
    \end{tabular}
\end{table}

Each agent-facing call (e.g., \texttt{create\_memory}, \texttt{llm\_chat})
is routed through the SDK to one or more of these module-level operations;
agents never call \texttt{Mem0Provider} or \texttt{MemoryWriteBarrier}
directly. This indirection is what allows the kernel to change providers,
retrieval strategies, or barrier timeouts without requiring changes to
agent code.

\subsection{Write Barrier Implementation}
\label{sec:appendix-barrier}

The write barrier described in Section~\ref{sec:write-barrier} is
implemented as a per-\texttt{user\_id} sequence counter with acquire and
release operations, sketched below:

\begin{quote}
\small
\texttt{
class MemoryWriteBarrier:\\
\ \ def acquire(self, user\_id):\\
\ \ \ \ \# assigns and returns the next sequence number for user\_id\\
\ \ \ \ ...\\
\\
\ \ def release(self, user\_id, seq\_no, success):\\
\ \ \ \ \# marks seq\_no as drained (committed or failed) for user\_id;\\
\ \ \ \ \# notifies any retrieval waiting on this or an earlier seq\_no\\
\ \ \ \ ...\\
\\
\ \ def snapshot(self, user\_id):\\
\ \ \ \ \# returns the current high-water mark $\sigma^*$(user\_id)\\
\ \ \ \ ...\\
\\
\ \ def wait\_until\_drained(self, user\_id, snapshot, timeout\_ms=5000):\\
\ \ \ \ \# blocks until all writes with seq\_no <= snapshot are released,\\
\ \ \ \ \# or returns early (fail-open) after timeout\_ms elapses\\
\ \ \ \ ...\\
}
\end{quote}

We choose a bounded wait with fail-open behavior over strict locking
because a stalled write degrades personalization rather than violating
correctness -- the assistant can still respond with partial context --
mirroring the fail-open/fail-closed asymmetry in
Section~\ref{sec:metadata-visibility} (liveness degrades gracefully;
visibility never fails open). A failed write still calls
\texttt{release} (with \texttt{success=False}), so a provider error never
strands a waiting retrieval past the bounded timeout. This cost is scoped
to a single \texttt{user\_id} and bounded by the number of writer agents
rather than memory-store size; we have not evaluated behavior at larger
memory-store sizes.

\subsection{Kernel-Managed Memory Injection}

The injection pipeline performs the following steps: extract the latest
user query; retrieve candidate memories using semantic search; derive the
target \texttt{user\_id} from retrieved metadata; perform cross-agent
retrieval for shared memories; merge and deduplicate results; filter by
relevance score; sort by relevance; format structured memory into natural
language; truncate to a token budget; and inject as a system message
prepended with \texttt{===== MEMORY CONTEXT =====}. This design ensures
personalization is applied uniformly without requiring agents to
explicitly request memory.

\subsection{Cross-Agent Memory Resolution}

The kernel resolves shared memory using a two-stage retrieval process:
retrieve agent-scoped memories, infer \texttt{user\_id}, then retrieve
shared memories from other agents. If no user ID is found locally, the
kernel falls back to a global registry of known users, enabling
cross-agent personalization even when the requesting agent has no prior
memory -- zero-shot personalization across agents and memory reuse
without explicit coordination.

\subsection{Evaluation Pipeline and Hybrid Metric}

The evaluation pipeline runs ProfileAgent, then TaskAgent, then
AssistantAgent under two conditions: Phase 1 (private memory only) and
Phase 2 (shared memory enabled), differing only in sharing policy. The
early hybrid scoring mechanism combined deterministic keyword matching
with LLM-based evaluation across Profile, Task, and Integration:
\begin{equation}
\text{score} = \frac{\text{keyword score} + \text{LLM score}}{2},
\end{equation}
trading robustness (keyword matching) for semantic evaluation (LLM
judge). Each trial in the synthetic evaluation harness logs injected
memory count, cross-agent retrieval, personalization scores, and latency,
aggregated into summary statistics including mean, standard deviation,
and min/max values.

\end{document}